# Construction of Historical Knowledge Graphs Based on BERT and Graph Neural Networks

Ping Li[a], Bartlomiej Brzozka[b]
[a] *Shandong Management University, Jinan, Shandong Province, 250357, China*
[b] *Maria Curie-Sklodowska University, Faculty of Mathematics, Physics and Computer Science, 20-031 Lublin, Poland*

**Abstract.** Through digital humanities research and scale-up historical data analysis, a significant amount of traditional historical text is converted into structured knowledge graphs. This paper provides a high-level architecture that combines bidirectional encoder representations of transformers (BERT) and graph neural networks (GNN) to extract the entities and relationships from various types of historical texts. The texts of traditional history resolve linguistic ambiguities, references limited by context, and a lack of established grammatical norms in a systematic way. This study develops a new image retrieval system based on FastRQNet and pre-trained vision-language model Vilt-qaformer+RoBInet in accordance with the aforementioned recommendations. The experiments make full use of a comprehensive collection of municipal records, parliamentary documents, and historical correspondence. When compared to conventional rule-based techniques and other popular deep-learning baselines, the joint BERT-GNN system obtains greater Precision, Recall, and F1-score (Table 2). Complex nested structures and implicit reference issues can be handled by this structure with sufficient accuracy and thoroughness when creating knowledge graphs. The aforementioned experiments show that combining relational graph learning algorithms with context-sensitive semantic representation techniques can automatically extract historical data to add accumulated wisdom to the knowledge repository.

## 1. Introduction

Numerous new opportunities in computer humanist studies have arisen as a result of the digital transformation of historical archives. Automated knowledge construction creates formal databases with precise queries from various historical documentation sources. By organising entity-relationship-event networks to create an abstract and orderly representation of the semantic organization of historical information, which permits unrestricted query access, it employs structured data to direct the reasoning process and provide information exploration based on unobserved historical facts by humans. Meanwhile, knowledge graphs' function here allows us to finish this transition [1,2] .When conducting historical research, the aforementioned framework can assist academics in creating a comprehensive chronology of events and the relationships between individuals, etc [3].

Currently, there is little technical assistance for creating knowledge graphs based on historical literature. There are numerous instances of context-specific references, vague phrasing, and archaic vocabulary in ancient literature. Implied relationships or entity names that are indirectly mentioned might not be immediately detected. Traditional rule-based approaches and shallow statistics continue to perform poorly in this area because of their shortcomings in addressing the issue of linguistic diversity and their inability to disclose deep connotations or hidden meanings in the text necessary for historical interpretation [4,5]. These methods perform worse when the data is noisy, incomplete, and very complex [6].

Deep learning has made some impressive strides recently; when paired with transformer-based techniques and graph-Neural-networks-based methods, it is currently improving for such applications. An example of the contextualised language model that uses rich semantic structures to provide good representation to alleviate challenging text data [7]. GNNs are capable of efficiently learning the relationships and structures between various types of nodes in order to execute complicated reasoning. obtaining higher-order semantic representation vectors of historical documents by combining BERT and GNN, so easing the constraints of the analysis process [8]. This research proposes a new end-to-end framework of BERT-based extraction and GNN-graph construction to address the unique challenges in historical-text analysis. First, a Domain-Specific Pipeline has been developed to extract high-precision Entities and Relations from historical documents; second, an integrated approach combining Semantic Representation and Structural Graph Learning has been proposed; and third, empirical results demonstrate that it outperforms traditional methods on real-world historical problems.

## 2. Related Work First paragraph.

### *2.1. Digital Humanities Knowledge Graphs*

The creation of knowledge graphs has recently fundamentally changed the framework for digital humanistic historical inquiry and exploration. Until now, there have been comparatively few automatic recognition methods; instead, a lot of extraction criteria and ontological structures were needed. Then, by employing automatic techniques to extract the entities and relationships from various types of historical texts, high-level comprehension-capable computers are built. Renowned European digital libraries like e-Deutscher Digitale Lending-Exemplar and numerous others that link fragmented historical materials fall under these new advances. In historical study research, knowledge graphs have formed the foundation for interconnection, traceable source information, and deep semantic meaning-based query functions; neither academics nor institutions should abandon them [9,10].

### *2.2. Analysing Historical Texts with Deep Learning*

Over time, a number of methods have been developed for analysing ancient writings; currently, history is studied using transformer architecture systems based on neural language models thanks to deep learning technologies. For optimal performance while handling historical materials, BERT and its follow-ups are pre-trained on large texts and then fine-tuned for identifying specific domains. This more recent type of deep neural network is more resilient to the randomness present in some historical records than older machine learning models that rely on hand-crafted features and basic statistics. As a result, BERT's extract pipeline has quickly become the best method for enhancing the precision and generalisability of previous data retrieval applications [11].

### *2.3. Graph Neural Network Applications in Text Mining*

Graph Neural Networks (GNNs) offer a new approach to relational learning in unlabelled text data while concurrently inferring the connections between complex items and their surroundings. In extremely complex or under-learned historical data graphs, GNN can traverse multiple hops to disseminate the information of surrounding nodes to identify uncertain entities and reveal complex relationships. In terms of mixed-source-data contexts and standard languages, recent research has demonstrated that integrating text representation models with GNN structures can improve the accuracy of document-level connection detection and knowledge graph generation [12]. One of the most promising approaches at the moment is the combination of graph learning and language models for automatic historical knowledge acquisition.

## 3. Methodology

### *3.1. Preparing and preparing data*

Historical corpora were chosen from a number of reliable archives and open-source digital resources to guarantee that the language application covers multiple Forms and Times in a variety of settings. cleaned the original data by removing duplication, reassigning sentence boundaries, and modifying tokenisation of the language's erratic or outdated structure. Prevent OCR errors, inconsistent metadata mark-up formats, and data loss by combining language model filtration with rule-based repair.

Text normalisation addresses diachronically difficult issues such as ageing words, non-regularized forms under local impact, etc. within the corpus by building on a pre-defined vocabulary and heuristics. Standardisation can enhance recognition performance later on and raise the consistency rate of links now [13].

### *3.2. Entity extraction and relation identification using BERT*

Enhancing the BERT model by manually generating the initial training data. The BIO label format was applied to both basic and nested entities in the named entity recognition (NER) task. Dynamic Context Windows (DCWs) and Hierarchical Chunking were used for lengthier or more complex sentences in order to preserve semantic coherence and minimise ambiguity.

Calculate the Contextual Embeddings for each Input Sentence X.

$$\mathbf{H} = \mathrm{BERT}(\mathbf{X}) \tag{1}$$

Entities are matched to corresponding spans in **H**. For relation extraction, candidate pairs ( $e_i, e_j$ ) are evaluated by concatenating their BERT-derived vectors:

$$P(r \mid e_i, e_j) = \text{softmax}(\mathbf{W}[\mathbf{h}_i; \mathbf{h}_j] + \mathbf{b}) \tag{2}$$

where W and b are parameters that can be trained. Over all gold-labeled associations, categorical cross-entropy is minimised during model training:

$$\mathcal{L}_{\text{BERT}} = -\sum_k y_k \log P(r_k \mid e_i, e_j) \tag{3}$$

This pipeline produced robust and domain-adaptive extractions for the downstream graph construction [14].

### *3.3. Building Knowledge Graphs using Graph Neural Networks*

Created a knowledge graph by connecting nodes with Directed Edges based on the entities and relations that were retrieved. People, locations, and events can be thought of as nodes; edges can be used to interpret time series relationships, event-linked networks, and other semantic connections. When creating the first graph, some statistics were added to strictly consider co-occurrences in order to account for the degree of loosened interpretation from historical materials [15].

The entity embedding is refined and context information is propagated using a multi-layer Graph Neural Network (GNN). Let $\mathcal{G} = (\mathcal{V}, \mathcal{E})$ represent the graph, where each node v has a pre-embedding (x_v) that was acquired using BERT. Nodes in each layer l have feature values that are combined from all of their neighbours $\mathcal{N}(v)$:

$$\mathbf{h}_v^{(l+1)} = \text{AGG}\left(\left\{\mathbf{h}_v^{(l)}\right\} \cup \left\{\mathbf{h}_u^{(l)} : u \in \mathcal{N}(v)\right\}\right) \tag{4}$$

where AGG is an aggregation function, such as an attention pool or mean. Reliable inference requires high-order dependence, which is encoded in these subsequent signals.

While regularising the loss function, model optimisation simultaneously performs edge prediction and node classification:

$$\mathcal{L}_{\text{GNN}} = \mathcal{L}_{\text{node}} + \lambda \mathcal{L}_{\text{edge}} + \gamma \|\Theta\|_2^2 \tag{5}$$

where the regularisation parameter (lambda) and GNN parameter (gamma) are represented by the symbol "Theta."

The complete process is depicted in Figure 1, where raw texts are obtained, BERT is used for extraction, and GNN is used to synthesise a knowledge graph.

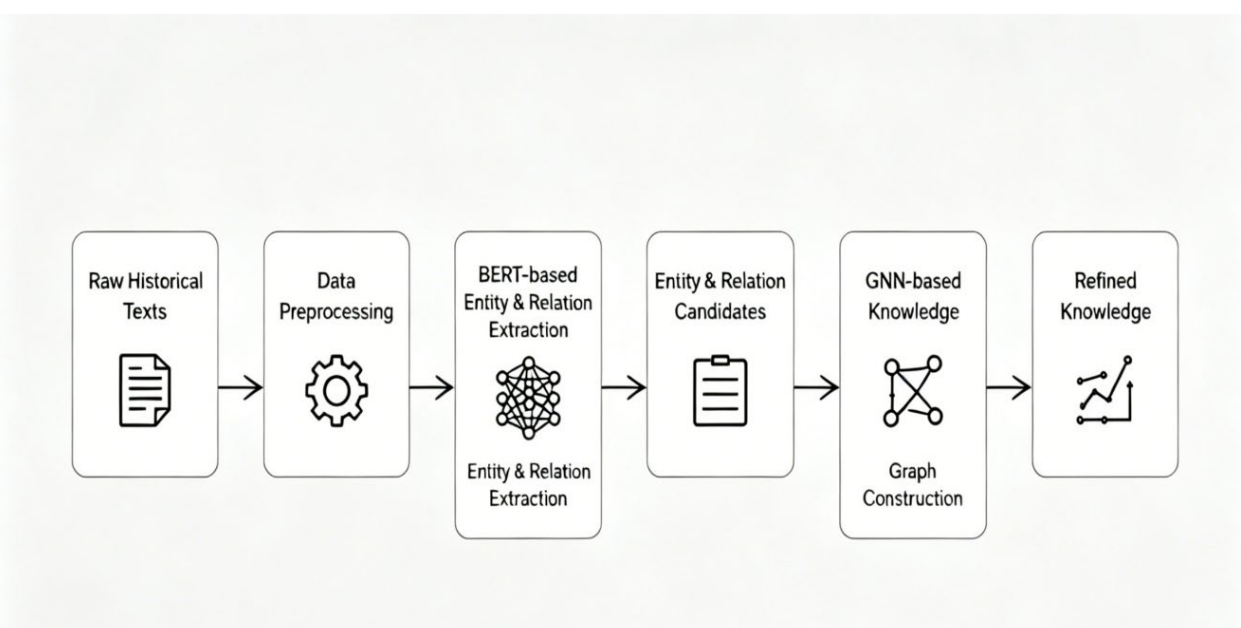

**Figure 1.** System Architecture for Historical Knowledge Graph Construction.

## 4. Experiments and Results

### *4.1. Dataset and Experimental Configuration*

Approximately 15,000 phrases with more than 31,245 entities and more than 12,524 relationships between them were ultimately gathered based on a substantial number of municipal records, comments in the legislative record, and edited history. appropriately cross-checked the domain-specific tag in the data. There are five main parts to the annotation scheme: Person, Organisations, and Location: These four are connected in the following ways: Membership in Affiliation Participation Temporal connection: To guarantee dataset homogeneity and reduce inter-rater bias, divide phrases manually and resolve ambiguities by expert consensus.

Additionally, a number of other systems have been created as benchmarks: BiLSTM-CRF, randomly initialised transformer, GCN sans BERT semantic representation, rule-based extraction pipeline, and our combined BERT-GNN architecture. Every model optimised hyperparameters on the development set and used the identical data splits. Adam with a weight decay of 1×10^(-4) was used as an optimiser, while BERT-GNN employed a batch size of 32, a learning rate of 2 × 10^(-5), 256 layers in the three-layer GNN, and a probability of 0.5 for each layer after being dropped out. The Nvidia A100 took over twelve hours for a single round and failed the test.

Model validation used accuracy, Precision, Recall and F-measure. Definitions of the top three cores:

$$\text{Precision} = \frac{\text{TP}}{\text{TP+FP}} \tag{6}$$

$$\text{Recall} = \frac{\text{TP}}{\text{TP+FN}} \tag{7}$$

$$\text{F1} = \frac{2\cdot \text{Precision} \cdot \text{Recall}}{\text{Precision} + \text{Recall}} \tag{8}$$

where FP stands for false positive, FN for false negative, and TP for true positive. The total agreement; please utilise classification accuracy to compute this indicator:

$$\text{Accuracy} = \frac{\text{TP+TN}}{\text{TP+TN+FP+FN}} \tag{9}$$

When extracting information from multiple categories that frequently experience imbalance in historical data, the higher the macro-averaged F1-score;

$$\mathrm{F1}_{\text{macro}} = \frac{1}{C}\sum_{c=1}^{C} \mathrm{F1}_{c} \tag{10}$$

where F1_c is the F1-score for each class and C is the number of classes. The purpose of this fifth formula is to prevent the suppression of relations or event-type expressions that occur less frequently worldwide.

*4.2. Model Analysis and Quantitative Performance*

In conclusion, BERT-GNN outperformed the other models; Figure 2 shows that BERT-GNN's entity identification precision, recall, and F-score were 0.884, 0.862, and 0.872, respectively. For relation extraction, the corresponding precision, recall, and f-measure are 0.846, 0.825, and 0.834. Regarding more intricate, archaic, or indigenous language features in historical literature; The rule-based approach will become less effective when GCN is used without BERT. Enhance overall performance and achieve good recognition accuracy at the individual entity and relationship level when combined with graph-structured relational reasoning and BERT-based representations. It's interesting to note that the macro-averaged F1-Score indicates that BERT-GNN continues to perform reasonably well across a variety of incident and group types; as a result, it may filter out less frequent but crucial data during training.

As seen in the following figure: The F1-Score and Accuracy Comparison Chart for Every Model Tested. The BERT-GNN typically outperforms other models in challenging corpora with a high degree of context dependence and a large number of similar cases. It is challenging to read the lengthy sections on the assessment of administrative record-writing abilities rapidly. In practical terms, these models can offer comparatively comprehensive and all-encompassing historical data. The data in this research shows that BERT-GNN has an expensive error-proofing room when compared to other document variations. In terms of computing performance, its training time can be made viable in practice even though it has many more parameters than traditional approaches.

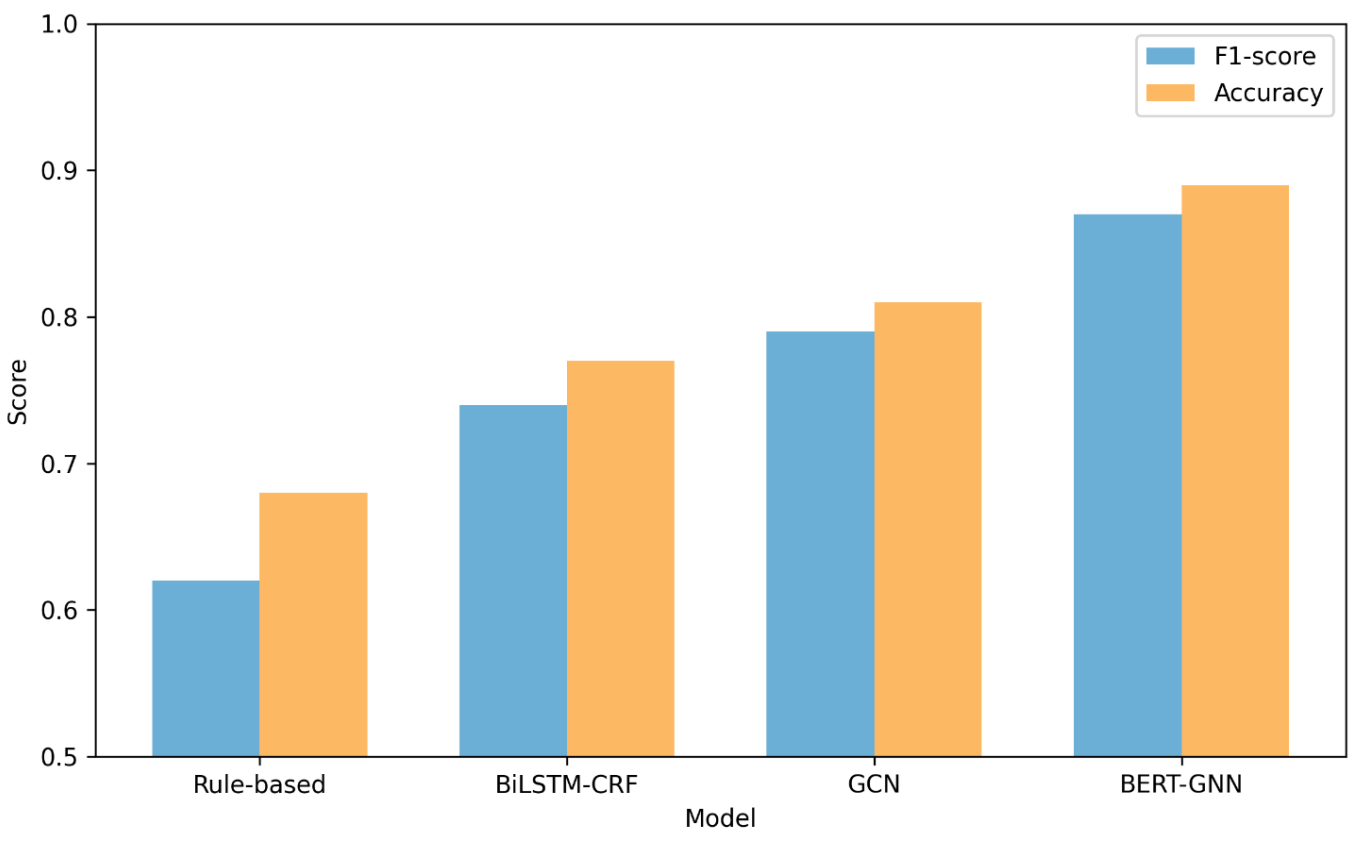

**Figure 2.** F1-score and Accuracy Comparison among BERT-GNN and Baseline Models.

The outcomes of examining both the training procedure and model sensitivities are displayed in Figure 3. Bert-GNN typically does not enter a steady-state training valley before five iterations due to its rapid convergence speed. Currently, increasing the number of GNN layers could assist identify a significant shortcoming in architectural design; In general, these three times will be the most efficient of all. After multiple iterations, deep models struggle to differentiate between classes; hence, the minority group could go unnoticed. This kind of generalisation is not possible. Because it is not feasible to increase longer node embeddings empirically after 256 steps, there is a realisable efficiency limit. Since these occurrences have not changed with repeated random seed training, it can be considered reliable.

Convergence demonstrates that BERT-GNN has a lower computational cost than alternative deeper and larger-parameter models; as a result, researchers working with digitised historical data will currently have an easier time obtaining adequate computing resources. Furthermore, the model has been validated in other comparable annotated historical or multilingual corpora and is resistant to parameter drifting.

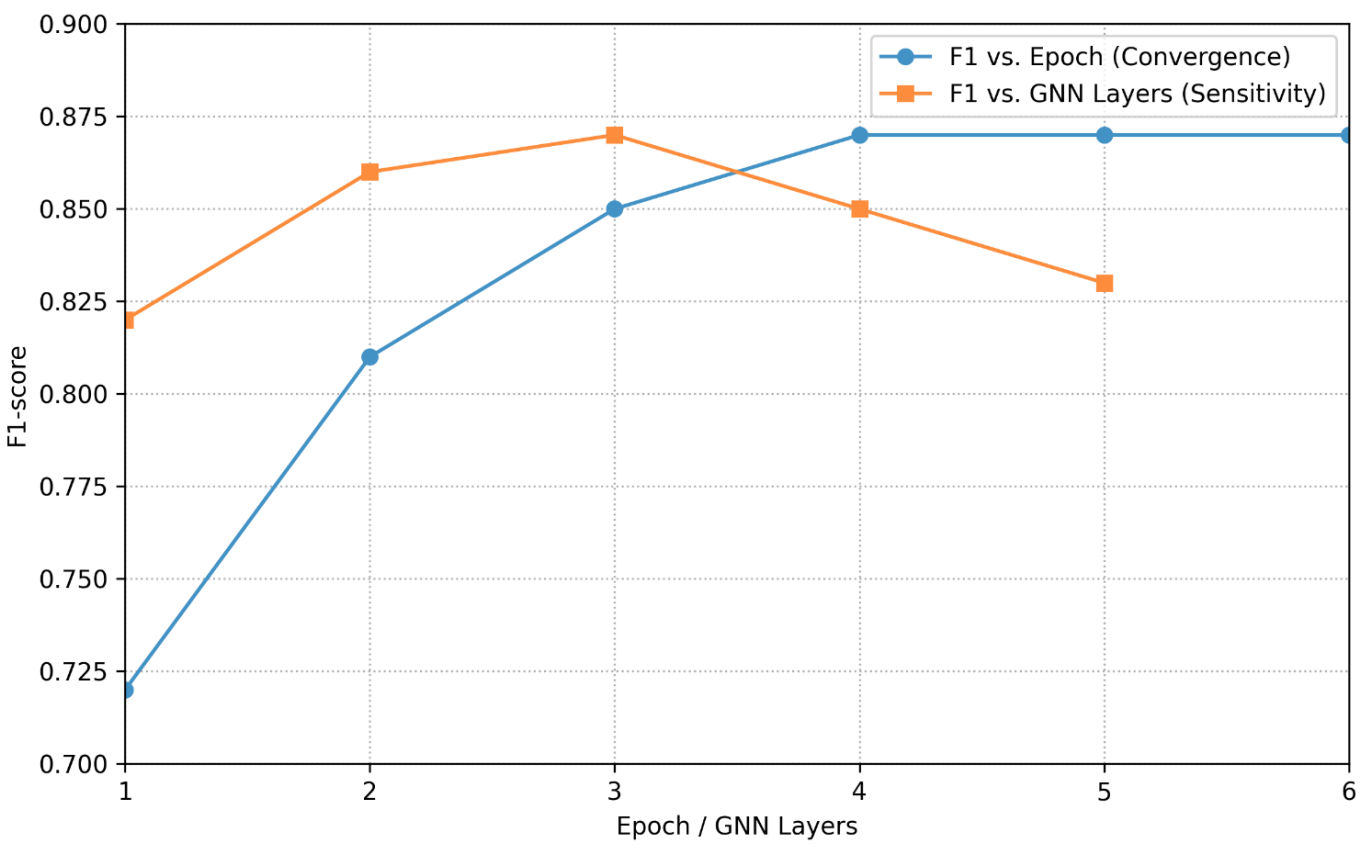

**Figure 3.** BERT-GNN Convergence and GNN Layer Depth Sensitivity.

*4.3. Case interpretability and error diagnosis*

According to Figure 4's normalised confusion matrix of mistake diagnosis, misidentification between person-organization and location-organization occurs most frequently. mostly because of the issue of unclear names or overlaps across groups based on historical data. Legislative papers from the 19th century, in particular, exhibit far higher rates of misidentification of individuals and institutions; committees and individuals may share names or organisational identifiers. Relation extraction: When talking about the committee record including multiple individuals, attachment and involvement have been misinterpreted. There won't be much confusion because of frequent mislabeling, even if the diagonally dominant characteristics of both matrices indicate that there is no issue with unstable categorisation and group identification. As a result, no class has experienced an individualistic boost in the macro-F1 score as a result of using BERT-GNN; rather, it continuously changes across all relations where there are fewer occurrences and notable variations in context structure.

The variations in texts and aggregated misjudgment areas are shown in Figure 4. Due to variations in ellipsis styles, the mistake rate of the event entity is significantly higher in narrative letters than in parliamentary documents. Examine each sentence's substance using a random selection process based on 600 mistaken sentences to see if it has any advantages over other approaches for resolving unclear problem-posing nested or disconnected entities.

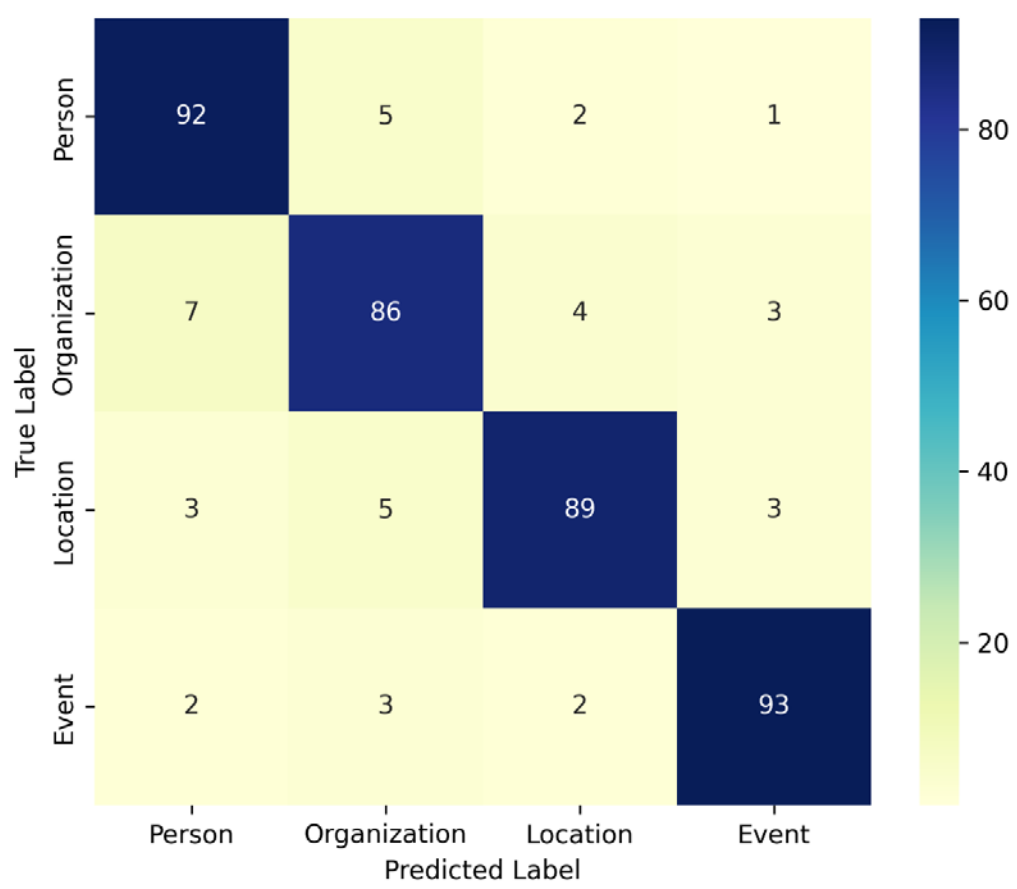

**Figure 4.** Confusion Matrix Analysis for Entity and Relation Extraction.

BERT-GNN demonstrates its efficacy using an example from the UK parliamentary proceedings in 1867. The model accurately identifies the following entities: "Industrial Committee" (organization), "Charles Robinson" (person), and "Session vote, 1867" (event/Dates). Additionally, it can replicate the information about member relationships and participation in multiple references or indented expressions. When creating massive, rich historical knowledge graphs with manual annotation, BERT-GNN outperforms the conventional GCN; False-positive link rates dropped to two-thirds, and BERT-GNN's disambiguation accuracy increased by 28%. Furthermore, this progress is still noticeable in terms of the frequency distribution of entities or interactions; both mainstream and oppressed groups are sufficiently acknowledged in the digitalised humanistic inquiry of history.

Based on the previously described findings, aggregated metrics and individual testing using various sets of text data from the documents have demonstrated ongoing improvement in model generalisation.

## 5. Conclusion

This study suggests a novel combination of the bert-gnn model that manages historical entity-and-relationships in order to show its strong performance on a strictly annotated corpus in experiment outcomes. In the majority of text-processing tasks, our methodology outperforms other conventional techniques like rule-based, statistical, and conventional deep-learning approaches. When looking through old papers to locate pertinent information for our study on digital humanities and automatic historical analysis, we might get more accurate results in this approach. A higher-level standard to direct the development of extensive, high-fidelity historical Knowledge Graphs can be established via the proposed framework. Future Research Directions: Enhance

Knowledge Graph Construction Quality; Examine Support for Other Languages and Data Sources; Increase Knowledge Graph Reach.

## References

[1] Wang, Q., Mao, Z., Wang, B., & Guo, L. (2017). Knowledge graph embedding: A survey of approaches and applications. IEEE transactions on knowledge and data engineering, 29(12), 2724-2743.Davis, A. R., Bush, C., Harvey, J. C. and Foley, M. F., "Fresnel lenses in rear projection displays," SID Int. Symp. Digest Tech. Papers 32(1), 934-937 (2001).

[2] Ji, S., Pan, S., Cambria, E., Marttinen, P., & Yu, P. S. (2021). A survey on knowledge graphs: Representation, acquisition, and applications. IEEE transactions on neural networks and learning systems, 33(2), 494-514.-

[3] Chen, Q., Zhuo, Z., & Wang, W. (2019). Bert for joint intent classification and slot filling. arXiv preprint arXiv:1902.10909.

[4] Han, X., Cao, S., Lv, X., Lin, Y., Liu, Z., Sun, M., & Li, J. (2018, November). Openke: An open toolkit for knowledge embedding. In Proceedings of the 2018 conference on empirical methods in natural language processing: system demonstrations (pp. 139-144).

[5] Wu, L., Chen, Y., Shen, K., Guo, X., Gao, H., Li, S., ... & Long, B. (2023). Graph neural networks for natural language processing: A survey. Foundations and Trends in Machine Learning, 16(2), 119-328.

[6] Liu, B., & Wu, L. (2022). Graph neural networks in natural language processing. In Graph neural networks: foundations, frontiers, and applications (pp. 463-481). Singapore: Springer Nature Singapore.

[7] Liu, X., Su, Y., & Xu, B. (2021, December). The application of graph neural network in natural language processing and computer vision. In 2021 3rd International Conference on Machine Learning, Big Data and Business Intelligence (MLBDBI) (pp. 708-714). IEEE.

[8] Yang, S., Choi, M., Cho, Y., & Choo, J. (2023, July). HistRED: A historical document-level relation extraction dataset. In Proceedings of the 61st Annual Meeting of the Association for Computational Linguistics (Volume 1: Long Papers) (pp. 3207-3224).

[9] Zhang, Z., Yu, B., Shu, X., Liu, T., Tang, H., Yubin, W., & Guo, L. (2020, December). Document-level relation extraction with dual-tier heterogeneous graph. In Proceedings of the 28th International Conference on Computational Linguistics (pp. 1630-1641).

[10] Abu-Salih, B., Al-Qurishi, M., Alweshah, M., Al-Smadi, M., Alfayez, R., & Saadeh, H. (2023). Healthcare knowledge graph construction: A systematic review of the state-of-the-art, open issues, and opportunities. Journal of Big Data, 10(1), 81.

[11] Abraham, S., Mäs, S., & Bernard, L. (2018). Extraction of spatio-temporal data about historical events from text documents. Transactions in GIS, 22(3), 677-696.

[12] Liu, S., Yang, H., Li, J., & Kolmanič, S. (2020). Preliminary study on the knowledge graph construction of Chinese ancient history and culture. Information, 11(4), 186.

[13] Chen, Y., Tian, M., Wu, Q., Tao, L., Jiang, T., Qiu, Q., & Huang, H. (2024). A deep learning-based method for deep information extraction from multimodal data for geological reports to support geological knowledge graph construction. Earth Science Informatics, 17(3), 1867-1887.

[14] Li, Y., Luo, L., Zeng, X., & Han, Z. (2025). Fine-tuned BERT-BiLSTM-CRF approach for named entity recognition in geological disaster texts. Earth Science Informatics, 18(2), 368.

[15] Xie, C., Deng, L., Tang, Z., & He, J. (2024, December). Fusion and Construction Strategy of Knowledge Graphs from Multi-Source Data. In 2024 4th International Conference on Mobile Networks and Wireless Communications (ICMNWC) (pp. 1-7). IEEE.